\documentclass[sigconf]{acmart} 

\usepackage{booktabs} 

\setcopyright{rightsretained} 

\copyrightyear{2019}
\acmYear{2019}
\setcopyright{acmlicensed}
\acmConference[GECCO '19 Companion]{Genetic and Evolutionary
Computation Conference Companion}{July 13--17, 2019}{Prague, Czech
Republic}
\acmBooktitle{Genetic and Evolutionary Computation Conference Companion
(GECCO '19 Companion), July 13--17, 2019, Prague, Czech Republic}
\acmPrice{15.00}
\acmDOI{10.1145/3319619.3321894}
\acmISBN{978-1-4503-6748-6/19/07}

\begin{document}
\title{AlphaStar: An Evolutionary Computation Perspective}

\author{Kai Arulkumaran}
\orcid{0000-0003-0459-892X}
\affiliation{%
  \institution{Imperial College London}
  \streetaddress{Exhibition Road}
  \city{London}
  \country{United Kingdom}
}
\email{ka709@ic.ac.uk}

\author{Antoine Cully}
\orcid{0000-0002-3190-7073}
\affiliation{%
  \institution{Imperial College London}
  \streetaddress{Exhibition Road}
  \city{London}
  \country{United Kingdom}
}
\email{a.cully@imperial.ac.uk}

\author{Julian Togelius}
\orcid{0000-0003-3128-4598}
\affiliation{%
  \institution{New York University}
  \streetaddress{2 MetroTech Center}
  \city{New York City}
  \state{NY}
  \country{United States}
}
\email{julian@togelius.com}

\renewcommand{\shortauthors}{K. Arulkumaran et al.}

\begin{abstract}
In January 2019, DeepMind revealed AlphaStar to the world---the first artificial intelligence (AI) system to beat a professional player at the game of StarCraft II---representing a milestone in the progress of AI. AlphaStar draws on many areas of AI research, including deep learning, reinforcement learning, game theory, and evolutionary computation (EC). In this paper we analyze AlphaStar primarily through the lens of EC, presenting a new look at the system and relating it to many concepts in the field. We highlight some of its most interesting aspects---the use of Lamarckian evolution, competitive co-evolution, and quality diversity. In doing so, we hope to provide a bridge between the wider EC community and one of the most significant AI systems developed in recent times.
\end{abstract}


\maketitle

\section{Background}
The field of artificial intelligence (AI) has long been involved in trying to create artificial systems that can rival humans in their intelligence, and as such, has looked to games as a way of challenging AI systems. Games are created by humans, for humans, and therefore have external validity to their use as AI benchmarks \cite{yannakakis2018artificial}.

After the defeat of the reigning chess world champion by Deep Blue in 1997, the next major milestone in AI versus human games was in 2016, when a Go grandmaster was defeated by AlphaGo \cite{silver2016mastering}. Both chess and Go were seen as some of the biggest challenges for AI, and arguably one of the few comparable tests remaining is to beat a grandmaster at StarCraft (SC), a real-time strategy game. Both the original game, and its sequel SC II, have several properties that make it considerably more challenging than even Go: real-time play, partial observability, no single dominant strategy, complex rules that make it hard to build a fast forward model, and a particularly large and varied action space.

DeepMind recently took a considerable step towards this grand challenge with AlphaStar, a neural-network-based AI system that was able to beat a professional SC II player in December 2018 \cite{alphastarblog}. This system, like its predecessor AlphaGo, was initially trained using imitation learning to mimic human play, and then improved through a combination of reinforcement learning (RL) and self-play. At this point the algorithms diverge, as AlphaStar utilises population-based training (PBT) \cite{jaderberg2017population} to explicitly keep a population of agents that train against each other \cite{jaderberg2018human}. This part of the training process was built upon multi-agent RL and game-theoretic perspectives \cite{lanctot2017unified,balduzzi2018re}, but the very notion of a population is central to evolutionary computation (EC), and hence we can examine AlphaStar through this lens as well\footnote{Note that we present a high-level overview of general interest, and have left aside the many deep links to the crossovers between EC and game theory \cite{smith1982evolution}.}.

\vspace{-1mm}
\section{Components}
\subsection{Lamarckian evolution}
Currently, the most popular approach to training the parameters of neural networks is backpropagation (BP). However, there are many methods to tune their hyperparameters, including evolutionary algorithms (EAs).  
A particularly synergistic approach is to use a memetic algorithm (MA), in which evolution is run as an outer optimisation algorithm, and individual solutions can be optimised by other means, such as BP, in an inner loop \cite{moscato1989evolution}. In this specific case, an MA can combine the exploration and global search properties of EAs with the efficient local search properties of BP.

PBT \cite{jaderberg2017population}, used in AlphaStar to train agents, is an MA that uses Lamarckian evolution (LE)\footnote{A more extensive literature review on LE can be found in the original paper.}: in the inner loop, neural networks are continuously trained using BP, while in the outer loop, networks are picked using one of several selection methods (such as binary tournament selection), with the winner's parameters overwriting the loser's; the loser also receives a mutated copy of the winner's hyperparameters \cite{goldberg1991comparative}. PBT was originally demonstrated on a range of supervised learning and RL tasks, tuning networks with higher performance than had previously been achieved. It is perhaps most beneficial in problems with highly non-stationary loss surfaces, such as deep RL, as it can change hyperparameters on the fly.

As a single network may take several gigabytes of memory, or need to train for several hours, scalability is key for PBT. As a consequence, PBT is both asynchronous and distributed \cite{nowostawski1999parallel}. Rather than running many experiments with static hyperparameters, the same amount of hardware can utilise PBT with little overhead---the outer loop reuses solution evaluation from the inner loop, and requires relatively little communication. When considering the effect of non-stationary hyperparameters and pre-emption on weaker solutions, the savings are even greater.

Another consequence of these requirements is that PBT is steady state \cite{syswerda1991study}, as opposed to generational EAs such as classic genetic algorithms. A natural fit for asynchronous EAs and LE, steady state EAs can allow the optimisation and evaluation of individual solutions to proceed uninterrupted and hence maximise resource efficiency. The fittest solutions survive longer, naturally providing a form of elitism/hall of fame, but even ancestors that aren't elites may be preserved, maintaining diversity\footnote{When given an appropriate selection pressure \cite{miller1995genetic}.}.

\vspace{-1mm}
\subsection{Co-evolution}
When optimising an agent to play a game, like in AlphaStar, it is possible to use self-play for the agent to improve itself. Competitive co-evolutionary algorithms (CCEAs) can be seen as a superset of self-play, as rather than keeping only a solution and its predecessors, it is instead possible to keep and evaluate against an entire population of solutions. Like self-play, CEAs form a natural curriculum \cite{hillis1990co}, but also confer an additional robustness as solutions are evaluated against a varied set of other solutions \cite{rosin1997new,stanley2004competitive}.

Through the use of PBT in a CCEA setting, Jaderberg et~al. \cite{jaderberg2018human} were able to train agents to play a first-person game from pixels, utilising BP-based deep RL in combination with evolved reward functions \cite{ackley1991interactions}. The design of CEAs have many aspects \cite{popovici2012coevolutionary}, and characterising this approach could lead to many potential variants. Here, for example, the interaction method was atypically based on sampling agents with similar fitness evaluations (Elo ratings), but many other heuristics exist.

\vspace{-1mm}
\subsection{Quality diversity}
A major advantage of keeping a population of solutions---as opposed to a single one---is that the population can represent a diverse set of solutions. This is not restricted strictly to multi-objective optimisation problems, but can also be applied to single objectives, where behaviour descriptors (BDs; i.e., solution phenotypes) can be used to pick solutions in the end. Quality diversity (QD) algorithms explicitly optimise for a single objective (quality), but also search for a large variety of solution types, via BDs, to encourage greater diversity in the population \cite{cully2018quality}. Recently, Ecoffet et~al. \cite{ecoffet2019go} used a QD algorithm to reach another milestone in playing games with AI---their system was the first to solve Montezuma's Revenge, a platform game notorious for its difficulty in exploring the environment.

In SC, there is no best strategy. Hence, the final AlphaStar agent consists of the set of solutions from the Nash distribution of the population---the set of complementary, least exploitable strategies \cite{balduzzi2018re}. In order to improve training, as well as increase the variety in the final set of solutions, it therefore makes sense to explicitly encourage diversity. As it does so, AlphaStar can also be classified as a QD algorithm. In particular, agents may have game-specific BDs, such as building extra units of a certain type, but also criteria to beat a certain other agent\footnote{A concept highly related to competitive fitness sharing in CCEAs \cite{rosin1997new}.}, criteria to beat a set of other agents, or even a mix of these. Furthermore, these specific criteria are also adapted online, which is relatively novel among QD algorithms \cite{wang2019paired}. There is more that could be done here though: it may be possible to extract BDs from human data \cite{yannakakis2018artificial}, or even learn them in an unsupervised manner \cite{cully2018hierarchical}. And, given a set of diverse strategies, a natural next step is to infer which might work best against a given opponent, enabling online adaptation.

\vspace{-1mm}
\section{Discussion}
While AlphaStar is a complex system that draws upon many areas of AI research, we believe a hitherto undersold perspective is that of it as an EA. In particular, it combines LE, CCEAs, and QD to spectacular effect. We hope that this perspective will give both the EC and deep RL communities the ability to better appreciate and build upon this significant AI system.

\bibliographystyle{ACM-Reference-Format}
\bibliography{references}


\begin{thebibliography}{22}


\ifx \showCODEN    \undefined \def \showCODEN     #1{\unskip}     \fi
\ifx \showDOI      \undefined \def \showDOI       #1{#1}\fi
\ifx \showISBNx    \undefined \def \showISBNx     #1{\unskip}     \fi
\ifx \showISBNxiii \undefined \def \showISBNxiii  #1{\unskip}     \fi
\ifx \showISSN     \undefined \def \showISSN      #1{\unskip}     \fi
\ifx \showLCCN     \undefined \def \showLCCN      #1{\unskip}     \fi
\ifx \shownote     \undefined \def \shownote      #1{#1}          \fi
\ifx \showarticletitle \undefined \def \showarticletitle #1{#1}   \fi
\ifx \showURL      \undefined \def \showURL       {\relax}        \fi
\providecommand\bibfield[2]{#2}
\providecommand\bibinfo[2]{#2}
\providecommand\natexlab[1]{#1}
\providecommand\showeprint[2][]{arXiv:#2}

\bibitem[\protect\citeauthoryear{Ackley and Littman}{Ackley and
  Littman}{1991}]%
        {ackley1991interactions}
\bibfield{author}{\bibinfo{person}{David Ackley} {and} \bibinfo{person}{Michael
  Littman}.} \bibinfo{year}{1991}\natexlab{}.
\newblock \showarticletitle{Interactions between learning and evolution}.
\newblock \bibinfo{journal}{{\em Artificial life II\/}}  \bibinfo{volume}{10}
  (\bibinfo{year}{1991}), \bibinfo{pages}{487--509}.
\newblock


\bibitem[\protect\citeauthoryear{Balduzzi, Tuyls, Perolat, and
  Graepel}{Balduzzi et~al\mbox{.}}{2018}]%
        {balduzzi2018re}
\bibfield{author}{\bibinfo{person}{David Balduzzi}, \bibinfo{person}{Karl
  Tuyls}, \bibinfo{person}{Julien Perolat}, {and} \bibinfo{person}{Thore
  Graepel}.} \bibinfo{year}{2018}\natexlab{}.
\newblock \showarticletitle{Re-evaluating evaluation}. In
  \bibinfo{booktitle}{{\em NeurIPS}}.
\newblock


\bibitem[\protect\citeauthoryear{Cully and Demiris}{Cully and Demiris}{2018a}]%
        {cully2018hierarchical}
\bibfield{author}{\bibinfo{person}{Antoine Cully} {and}
  \bibinfo{person}{Yiannis Demiris}.} \bibinfo{year}{2018}\natexlab{a}.
\newblock \showarticletitle{Hierarchical Behavioral Repertoires with
  Unsupervised Descriptors}. In \bibinfo{booktitle}{{\em GECCO}}.
\newblock


\bibitem[\protect\citeauthoryear{Cully and Demiris}{Cully and Demiris}{2018b}]%
        {cully2018quality}
\bibfield{author}{\bibinfo{person}{Antoine Cully} {and}
  \bibinfo{person}{Yiannis Demiris}.} \bibinfo{year}{2018}\natexlab{b}.
\newblock \showarticletitle{Quality and diversity optimization: A unifying
  modular framework}.
\newblock \bibinfo{journal}{{\em TEVC\/}} \bibinfo{volume}{22},
  \bibinfo{number}{2} (\bibinfo{year}{2018}), \bibinfo{pages}{245--259}.
\newblock


\bibitem[\protect\citeauthoryear{Ecoffet, Huizinga, Lehman, Stanley, and
  Clune}{Ecoffet et~al\mbox{.}}{2019}]%
        {ecoffet2019go}
\bibfield{author}{\bibinfo{person}{Adrien Ecoffet}, \bibinfo{person}{Joost
  Huizinga}, \bibinfo{person}{Joel Lehman}, \bibinfo{person}{Kenneth~O
  Stanley}, {and} \bibinfo{person}{Jeff Clune}.}
  \bibinfo{year}{2019}\natexlab{}.
\newblock \showarticletitle{Go-Explore: a New Approach for Hard-Exploration
  Problems}.
\newblock \bibinfo{journal}{{\em arXiv preprint arXiv:1901.10995\/}}
  (\bibinfo{year}{2019}).
\newblock


\bibitem[\protect\citeauthoryear{Goldberg and Deb}{Goldberg and Deb}{1991}]%
        {goldberg1991comparative}
\bibfield{author}{\bibinfo{person}{David~E Goldberg} {and}
  \bibinfo{person}{Kalyanmoy Deb}.} \bibinfo{year}{1991}\natexlab{}.
\newblock \showarticletitle{A comparative analysis of selection schemes used in
  genetic algorithms}.
\newblock In \bibinfo{booktitle}{{\em Foundations of Genetic Algorithms}}.
  Vol.~\bibinfo{volume}{1}. \bibinfo{pages}{69--93}.
\newblock


\bibitem[\protect\citeauthoryear{Hillis}{Hillis}{1990}]%
        {hillis1990co}
\bibfield{author}{\bibinfo{person}{W~Daniel Hillis}.}
  \bibinfo{year}{1990}\natexlab{}.
\newblock \showarticletitle{Co-evolving parasites improve simulated evolution
  as an optimization procedure}.
\newblock \bibinfo{journal}{{\em Physica D: Nonlinear Phenomena\/}}
  \bibinfo{volume}{42}, \bibinfo{number}{1-3} (\bibinfo{year}{1990}),
  \bibinfo{pages}{228--234}.
\newblock


\bibitem[\protect\citeauthoryear{Jaderberg, Czarnecki, Dunning, Marris, Lever,
  et~al\mbox{.}}{Jaderberg et~al\mbox{.}}{2018}]%
        {jaderberg2018human}
\bibfield{author}{\bibinfo{person}{Max Jaderberg}, \bibinfo{person}{Wojciech~M
  Czarnecki}, \bibinfo{person}{Iain Dunning}, \bibinfo{person}{Luke Marris},
  \bibinfo{person}{Guy Lever}, {et~al\mbox{.}}}
  \bibinfo{year}{2018}\natexlab{}.
\newblock \showarticletitle{Human-level performance in first-person multiplayer
  games with population-based deep reinforcement learning}.
\newblock \bibinfo{journal}{{\em arXiv preprint arXiv:1807.01281\/}}
  (\bibinfo{year}{2018}).
\newblock


\bibitem[\protect\citeauthoryear{Jaderberg, Dalibard, Osindero, Czarnecki,
  Donahue, et~al\mbox{.}}{Jaderberg et~al\mbox{.}}{2017}]%
        {jaderberg2017population}
\bibfield{author}{\bibinfo{person}{Max Jaderberg}, \bibinfo{person}{Valentin
  Dalibard}, \bibinfo{person}{Simon Osindero}, \bibinfo{person}{Wojciech~M
  Czarnecki}, \bibinfo{person}{Jeff Donahue}, {et~al\mbox{.}}}
  \bibinfo{year}{2017}\natexlab{}.
\newblock \showarticletitle{Population based training of neural networks}.
\newblock \bibinfo{journal}{{\em arXiv preprint arXiv:1711.09846\/}}
  (\bibinfo{year}{2017}).
\newblock


\bibitem[\protect\citeauthoryear{Lanctot, Zambaldi, Gruslys, Lazaridou, Tuyls,
  et~al\mbox{.}}{Lanctot et~al\mbox{.}}{2017}]%
        {lanctot2017unified}
\bibfield{author}{\bibinfo{person}{Marc Lanctot}, \bibinfo{person}{Vinicius
  Zambaldi}, \bibinfo{person}{Audrunas Gruslys}, \bibinfo{person}{Angeliki
  Lazaridou}, \bibinfo{person}{Karl Tuyls}, {et~al\mbox{.}}}
  \bibinfo{year}{2017}\natexlab{}.
\newblock \showarticletitle{A unified game-theoretic approach to multiagent
  reinforcement learning}. In \bibinfo{booktitle}{{\em NeurIPS}}.
  \bibinfo{pages}{4190--4203}.
\newblock


\bibitem[\protect\citeauthoryear{Miller and Goldberg}{Miller and
  Goldberg}{1995}]%
        {miller1995genetic}
\bibfield{author}{\bibinfo{person}{Brad~L Miller} {and}
  \bibinfo{person}{David~E Goldberg}.} \bibinfo{year}{1995}\natexlab{}.
\newblock \showarticletitle{Genetic algorithms, tournament selection, and the
  effects of noise}.
\newblock \bibinfo{journal}{{\em Complex Systems\/}} \bibinfo{volume}{9},
  \bibinfo{number}{3} (\bibinfo{year}{1995}), \bibinfo{pages}{193--212}.
\newblock


\bibitem[\protect\citeauthoryear{Moscato}{Moscato}{1989}]%
        {moscato1989evolution}
\bibfield{author}{\bibinfo{person}{P Moscato}.}
  \bibinfo{year}{1989}\natexlab{}.
\newblock \bibinfo{booktitle}{{\em On Evolution, Search, Optimization, Genetic
  Algorithms and Martial Arts: Towards Memetic Algorithms}}.
\newblock \bibinfo{type}{{T}echnical {R}eport}.
  \bibinfo{institution}{California Institute of Technology}.
\newblock


\bibitem[\protect\citeauthoryear{Nowostawski and Poli}{Nowostawski and
  Poli}{1999}]%
        {nowostawski1999parallel}
\bibfield{author}{\bibinfo{person}{Mariusz Nowostawski} {and}
  \bibinfo{person}{Riccardo Poli}.} \bibinfo{year}{1999}\natexlab{}.
\newblock \showarticletitle{Parallel genetic algorithm taxonomy}. In
  \bibinfo{booktitle}{{\em KES}}. \bibinfo{pages}{88--92}.
\newblock


\bibitem[\protect\citeauthoryear{Popovici, Bucci, Wiegand, and
  De~Jong}{Popovici et~al\mbox{.}}{2012}]%
        {popovici2012coevolutionary}
\bibfield{author}{\bibinfo{person}{Elena Popovici}, \bibinfo{person}{Anthony
  Bucci}, \bibinfo{person}{R~Paul Wiegand}, {and} \bibinfo{person}{Edwin~D
  De~Jong}.} \bibinfo{year}{2012}\natexlab{}.
\newblock \showarticletitle{Coevolutionary principles}.
\newblock In \bibinfo{booktitle}{{\em Handbook of Natural Computing}}.
  \bibinfo{pages}{987--1033}.
\newblock


\bibitem[\protect\citeauthoryear{Rosin and Belew}{Rosin and Belew}{1997}]%
        {rosin1997new}
\bibfield{author}{\bibinfo{person}{Christopher~D Rosin} {and}
  \bibinfo{person}{Richard~K Belew}.} \bibinfo{year}{1997}\natexlab{}.
\newblock \showarticletitle{New methods for competitive coevolution}.
\newblock \bibinfo{journal}{{\em Evolutionary Computation\/}}
  \bibinfo{volume}{5}, \bibinfo{number}{1} (\bibinfo{year}{1997}),
  \bibinfo{pages}{1--29}.
\newblock


\bibitem[\protect\citeauthoryear{Silver, Huang, Maddison, Guez, Sifre,
  et~al\mbox{.}}{Silver et~al\mbox{.}}{2016}]%
        {silver2016mastering}
\bibfield{author}{\bibinfo{person}{David Silver}, \bibinfo{person}{Aja Huang},
  \bibinfo{person}{Chris~J Maddison}, \bibinfo{person}{Arthur Guez},
  \bibinfo{person}{Laurent Sifre}, {et~al\mbox{.}}}
  \bibinfo{year}{2016}\natexlab{}.
\newblock \showarticletitle{Mastering the game of Go with deep neural networks
  and tree search}.
\newblock \bibinfo{journal}{{\em Nature\/}} \bibinfo{volume}{529},
  \bibinfo{number}{7587} (\bibinfo{year}{2016}), \bibinfo{pages}{484}.
\newblock


\bibitem[\protect\citeauthoryear{Smith}{Smith}{1982}]%
        {smith1982evolution}
\bibfield{author}{\bibinfo{person}{John~Maynard Smith}.}
  \bibinfo{year}{1982}\natexlab{}.
\newblock \bibinfo{booktitle}{{\em Evolution and the Theory of Games}}.
\newblock \bibinfo{publisher}{Cambridge University Press}.
\newblock


\bibitem[\protect\citeauthoryear{Stanley and Miikkulainen}{Stanley and
  Miikkulainen}{2004}]%
        {stanley2004competitive}
\bibfield{author}{\bibinfo{person}{Kenneth~O Stanley} {and}
  \bibinfo{person}{Risto Miikkulainen}.} \bibinfo{year}{2004}\natexlab{}.
\newblock \showarticletitle{Competitive coevolution through evolutionary
  complexification}.
\newblock \bibinfo{journal}{{\em JAIR\/}}  \bibinfo{volume}{21}
  (\bibinfo{year}{2004}), \bibinfo{pages}{63--100}.
\newblock


\bibitem[\protect\citeauthoryear{Syswerda}{Syswerda}{1991}]%
        {syswerda1991study}
\bibfield{author}{\bibinfo{person}{Gilbert Syswerda}.}
  \bibinfo{year}{1991}\natexlab{}.
\newblock \showarticletitle{A study of reproduction in generational and
  steady-state genetic algorithms}.
\newblock In \bibinfo{booktitle}{{\em Foundations of Genetic Algorithms}}.
  Vol.~\bibinfo{volume}{1}. \bibinfo{pages}{94--101}.
\newblock


\bibitem[\protect\citeauthoryear{Vinyals, Babuschkin, Chung, Mathieu,
  Jaderberg, et~al\mbox{.}}{Vinyals et~al\mbox{.}}{2019}]%
        {alphastarblog}
\bibfield{author}{\bibinfo{person}{Oriol Vinyals}, \bibinfo{person}{Igor
  Babuschkin}, \bibinfo{person}{Junyoung Chung}, \bibinfo{person}{Michael
  Mathieu}, \bibinfo{person}{Max Jaderberg}, {et~al\mbox{.}}}
  \bibinfo{year}{2019}\natexlab{}.
\newblock \bibinfo{title}{{AlphaStar: Mastering the Real-Time Strategy Game
  StarCraft II}}.
\newblock
  \bibinfo{howpublished}{\url{https://deepmind.com/blog/alphastar-mastering-real-time-strategy-game-starcraft-ii/}}.
    (\bibinfo{year}{2019}).
\newblock


\bibitem[\protect\citeauthoryear{Wang, Lehman, Clune, and Stanley}{Wang
  et~al\mbox{.}}{2019}]%
        {wang2019paired}
\bibfield{author}{\bibinfo{person}{Rui Wang}, \bibinfo{person}{Joel Lehman},
  \bibinfo{person}{Jeff Clune}, {and} \bibinfo{person}{Kenneth~O Stanley}.}
  \bibinfo{year}{2019}\natexlab{}.
\newblock \showarticletitle{Paired Open-Ended Trailblazer (POET): Endlessly
  Generating Increasingly Complex and Diverse Learning Environments and Their
  Solutions}.
\newblock \bibinfo{journal}{{\em arXiv preprint arXiv:1901.01753\/}}
  (\bibinfo{year}{2019}).
\newblock


\bibitem[\protect\citeauthoryear{Yannakakis and Togelius}{Yannakakis and
  Togelius}{2018}]%
        {yannakakis2018artificial}
\bibfield{author}{\bibinfo{person}{Georgios~N Yannakakis} {and}
  \bibinfo{person}{Julian Togelius}.} \bibinfo{year}{2018}\natexlab{}.
\newblock \bibinfo{booktitle}{{\em Artificial Intelligence and Games}}.
\newblock \bibinfo{publisher}{Springer}.
\newblock


\end{thebibliography}
\end{document}